\documentclass[journa]{IEEEtran}
\usepackage[dvips]{graphicx}
\usepackage{amssymb,amsmath,array}
\usepackage{psfrag} 
\usepackage{amsmath,amsthm} 
\usepackage[x11names]{xcolor}
\usepackage{overpic}
\usepackage{multirow}
\usepackage{rotating}
\usepackage{overpic}
\usepackage{makecell}
\usepackage{enumerate}

\usepackage{array} 
\usepackage{graphicx} 
\usepackage{subfigure}
\usepackage{caption}    
\usepackage{color}
\newcommand{\cb}[1]{\ifmmode {\boldsymbol{#1}}\else ${\boldsymbol{#1}}$\fi}
\newcommand{\cp}[1]{\ifmmode {\mathcal{#1}}\else ${\mathcal{#1}}$\fi}

\newcommand{\bx}{\cb{x}}
\newcommand{\bz}{\cb{z}}

\newcommand{\bX}{\cb{X}}
\newcommand{\bE}{\cb{E}}
\newcommand{\be}{\cb{e}}
\newcommand{\bA}{\cb{A}}

\begin{document}
%
\title{Bi-Objective Nonnegative Matrix Factorization: \\Linear Versus Kernel-Based Models}
%
%
%

\author{Fei~Zhu,
        Paul~Honeine,~\IEEEmembership{Member,~IEEE}
\thanks{F. Zhu and P. Honeine are with the Institut Charles Delaunay (CNRS), Universit\'{e} de Technologie de Troyes, France.\protect\\
Email : fei.zhu@utt.fr and paul.honeine@utt.fr}


}

\maketitle

\begin{abstract}
Nonnegative matrix factorization (NMF) is a powerful class of feature extraction 
 techniques that has been successfully applied in many fields, namely in signal and image processing. Current NMF techniques have been limited to a single-objective problem in either its linear or nonlinear kernel-based formulation. In this paper, we propose to revisit the NMF as a multi-objective problem, in particular a bi-objective one, where the objective functions defined in both input and feature spaces are taken into account. By taking the advantage of the sum-weighted method from the literature of multi-objective optimization, the proposed bi-objective NMF determines a set of nondominated, Pareto optimal, solutions instead of a single optimal decomposition. Moreover, the corresponding Pareto front is studied and approximated. Experimental results on unmixing real hyperspectral images confirm the efficiency of the proposed bi-objective NMF compared with the state-of-the-art methods.

\end{abstract}


%
\IEEEpeerreviewmaketitle

\section{Introduction}
%
%
%
%
\bigskip\bigskip

\IEEEPARstart{N}{onnegative matrix factorization} (NMF) provides a parts-based representation for the nonnegative data entries, and has becoming a versatile technique with plenty of applications \cite{lee99}. As opposed to other dimensionality reduction approaches, {\em e.g.,} principal component analysis, vector quantization and linear discriminant analysis, 
the NMF is based on the additivity of the contributions of the bases to approximate the original data. Such decomposition model often yields a tractable physical interpretation thanks to the sparse and nonnegative obtained representation of the input data. Many real world applications benefit from these virtues, including hyperspectral unmixing \cite{Jia2009,Fei14}, face and facial expression recognition \cite{polyNMF,GraphNMF}, gene expression data \cite{kim03}, 
 blind source separation \cite{cichocki2006new}, and spectral clustering \cite{Li2006,Ding2005}, to name a few.

The NMF approximates a high-rank nonnegative input matrix by two nonnegative low-rank ones. As a consequence, it provides a decomposition suitable for many signal processing and data analysis problems, and in particular the hyperspectral unmixing problem. Indeed, a hyperspectral image is a cube that consists of a set of images of the scene under scrutiny, each corresponding to a ground scene from which the light of certain wavelength
is reflected. 
Namely, a reflectance spectral over a wavelength range is available for each pixel. It is assumed that each spectral is a mixture of a few ``pure'' materials, called endmembers. The hyperspectral unmixing problem consists of extracting the endmembers (recorded in the first low-rank matrix), and estimating the abundance of each endmember at every pixel (recorded in the second one). Obviously, the above physical interpretation requires the nonnegativity on both abundances and endmember spectrums.

The NMF is a linear model, since it can be viewed in a way that each input spectral is approximated by a linear combination of some basis spectrums. To estimate the decomposition, the objective function for minimization is defined in an Euclidean space --- the so-called \emph{input space} \cp{X} ---, where the difference between the input matrix and the product of the estimated ones is usually measured either by the Frobenius norm or by generalized Kullback-Leibler divergence \cite{lee00}. These objective functions are often augmented by including different regularization terms, such as the Fisher constraint for learning local features \cite{Wang04fisher}, the sparseness constraint for intuitive and easily interpretable decompositions \cite{Hoyer04}, the temporal smoothness and spatial decorrelation regularization \cite{Cichocki}, and the minimum dispersion regularization for unmixing accuracy \cite{Huck}. Other objective functions are also raised from practical standpoints, {\em e.g.,} the $\ell_1$-norm for the robustness against outliers and missing data \cite{ke2005robust} 
and the Bregman divergence with fast computational performance \cite{Li12BregmanDivergence}. 

Many studies have shown the limits of a linear decomposition, as opposed to a nonlinear one \cite{14.tgrs.nonlinear,13.tsp.unmix,13.hype.nguyen}. While most research activities have been concentrated on the linear NMF, a few works have considered the nonlinear case. In an attempt to extent the linear NMF models to the nonlinear scope, several kernel-based NMF have been proposed within the framework offered by the kernel machines \cite{Zhang2006}. Employing a nonlinear function, the kernel-based methods mainly map the data into a higher dimensional space, where the existing linear techniques are performed on the transformed data. The kernel trick enables the estimation of the inner product between any pair of mapped data in a reproducing kernel Hilbert space --- the so-called \emph{feature space} \cp{H} ---, without the need of knowing explicitly neither the nonlinear map function nor the resulting space. For example, in \cite{Zhang2006}, the linear NMF technique is performed on the kernel matrix, whose entries consist of the inner products between input data calculated with some kernel function. Other kernel-based NMF techniques presented in \cite{Ding10,Li2012,Lee2009} follow a similar scheme but share an additive assumption originated from the convex NMF approach proposed in \cite{Ding10}, that is, the basis matrix is represented as the convex combination of the mapped input data in the feature space \cp{H}. It is worth noting that the objective function is the Frobenius norm of the residual between the kernel matrix and its factorization, for all the above-mentioned kernel-based NMF methods. However, although the input data matrix is nonnegative, the nonnegativity of the mapped data is not guaranteed. A more severe disadvantage is that the obtained bases lie in the feature space (often of infinite dimension), where a reverse mapping to the input space is difficult. Indeed, one needs to solve the pre-image problem, an obstacle inherited from the kernel machines \cite{11.spm}. In \cite{Fei14}, these difficulties are circumvented by defining a model in the feature space that can be optimized directly in the input space. 
In this paper, we revisit this framework to discover the nonlinearity of the input matrix. See Section~\ref{subsec:KNMF} for more details.

In either its linear conventional formulation or its nonlinear kernel-based formulation, as well as all of their variations (and regularizations), the NMF has been tackling a single-objective optimization problem. In essence, the underlying assumption is that it is known in prior that the linear model dominates the nonlinear one, or vice versa, for the input data under study. To obtain such prior information about the given input data is not practical in most real-world applications. Moreover, it is possible that a fusion of the linear and nonlinear models reveals the latent variables closer to the ground truth than each single model considered alone. Independently from the NMF framework, such combination of the linear model with a nonlinear fluctuation was recently studied by Chen, Richard and Honeine in \cite{13.tsp.unmix} and \cite{13.whispers.postnonlinear} where, in the former, the nonlinearity depends only on the spectral content, while it is defined by a post-nonlinear model in the latter. A multiple-kernel learning approach was studied in \cite{12.whispers.nonlinear} and a Bayesian approach was investigated in \cite{AltmannDMT14} with the so-called residual component analysis. All these methods share one major drawback: they only consist in estimating the abundances, with a nonlinear model, while the endmembers need to be extracted in a pre-processing stage using any conventional linear technique (N-Findr, vertex component analysis, ... \cite{VCA}). As opposed to such separation in the optimization problems, the NMF provides an elegant framework for solving jointly the unmixing problem, namely estimating the endmembers and the abundances. To the best of our knowledge, there have been no previous studies that combine the linear and nonlinear models within the NMF framework.


In this paper, we study the bi-objective optimization problem that performs simultaneously the NMF in both input and feature spaces, by combining the linear and kernel-based models. The first objective function to optimize stems from the conventional linear NMF, while the second objective function, defined in the feature space, is derived from a kernel-based NMF model. In case of two conflicting objective functions, there exists rather a set of nondominated, noninferior or Pareto optimal solutions, as opposed to the unique decomposition when dealing exclusively with one objective function. In order to acquire the Pareto optimal solutions, we investigate the sum-weighed method from the literature of multi-objective optimization, due to its ease for being integrated to the proposed framework. Moreover, propose to approximate the corresponding Pareto front. The multiplicative update rules are derived for the resulting sub-optimization problem in the case where the feature space is induced by the Gaussian kernel.  The convergence of the algorithm is discussed, as well as initialization and stopping criteria.

The remainder of this paper is organized as follows. We first revisit the conventional and kernel-based NMF. The differences between the input and the feature space optimization are discussed in Section~\ref{Sec:inputVSfeature}. In Section~\ref{Sec:ProposedMethod}, we present the proposed bi-objective NMF framework. Section~\ref{Sec:Experiments} demonstrates the efficiency of the proposed method for unmixing two real hyperspectral images. Conclusions and future works are reported in Section~\ref{Sec:Conclusion}.

\section{A primer on the linear and nonlinear NMF}

In this section, we present the two NMF variants, with the linear and the nonlinear models, as well as the corresponding optimization problems.

\subsection{Conventional NMF}\label{subsec:Conventional NMF}

Given a nonnegative data matrix $\bX
\in \Re^{L\times T}$, the conventional NMF aims to approximate it by the product of two low-rank nonnegative matrices $\bE
\in \Re^{L\times N}$ and $\bA \in \Re^{N\times T}$, 
 namely
\begin{equation}\label{eq:NMF0}
	\bX \approx \bE \bA,
\end{equation}
under the constraints $\bE \geq 0$ and $\bA \geq 0$, where the nonnegativity is element-wise. An equivalent vector-wise model is given by considering separately each column of the matrix $\bX$, namely $\bx_t$ for $t=1,\ldots,T$, with
\begin{equation*}
    \bx_t \approx \sum_{n=1}^N a_{nt} \, \be_n,
\end{equation*}
where each $\bx_t$ is represented as a linear combination of the columns of $\bE$, denoted $\be_n$ for $n=1,\ldots,N$, with the scalars $a_{nt}$ for $n=1,\ldots,N$ and $t=1,\ldots, T$ being the entries of the matrix $\bA$. The subspace spanned by the vectors $\bx_t$, as well as the vectors $\be_n$ is denoted the input space $\cp{X}$.

To estimate both matrices $\bE$ and $\bA$, one concentrates on the minimization of the Frobenius squared error norm $\frac12 \|\bX - \bE \bA\|^2_F$, subject to $\bE \geq 0$ and $\bA \geq 0$. In its vector-wise formulation, the objective function to minimize is
\begin{equation}\label{eq:NMF}
  J_\cp{X}(\bE,\bA)= \frac12 \sum_{t=1}^T\|\bx_t - \sum_{n=1}^N a_{nt} \, \be_n\|^2,
\end{equation}
where the residual error is measured between each input vector $\bx_t$ and its approximation $\sum_{n=1}^N a_{nt} \be_n$ in the input space $\cp{X}$. The optimization is operated with a two-block coordinate descent scheme, by alternating between the elements of $\bE$ or of $\bA$, while keeping the elements in the other matrix fixed.

\subsection{Nonlinear -- kernel-based -- NMF}\label{subsec:KNMF}

A straightforward generalization to the nonlinear form is proposed within the framework offered by the kernel machines. In the following, we present the kernel-based NMF that we have recently proposed in \cite{Fei14}. It is worth noting that other variants can also be investigated, including the ones studied in \cite{Zhang2006,Ding10,Li2012,Lee2009}. However, these variants suffer from the pre-image problem, making the derivations and the study more difficult; see \cite{15.tpami.knmf} for more details.

Consider a nonlinear function $\Phi(\cdot)$ that maps the columns of the matrix $\bX$, as well as the columns of the matrix $\bE$, from the input space $\cp{X}$ to some feature space $\cp{H}$. Its associated norm is denoted $\| \,\cdot\, \|_\cp{H}$, and the corresponding inner product in the feature space is of the form $\langle \Phi(\bx_{t}) , \Phi(\bx_{t'}) \rangle_\cp{H}$, which can be evaluated using the so-called kernel function $\kappa(\bx_{t},\bx_{t'})$ in kernel machines. Examples of kernel functions are the Gaussian and the polynomial kernels.

Applying the model \eqref{eq:NMF0} in the feature space, we get
 the following matrix factorization model
\begin{equation*}
    [\Phi(\bx_1) ~~ \Phi(\bx_2) ~ \cdots ~ \Phi(\bx_T)] \approx [\Phi(\be_1) ~~ \Phi(\be_2) ~ \cdots ~ \Phi(\be_N)] \bA,
\end{equation*}
or equivalently in the vector-wise form, for all $t=1,\ldots,T$,
\begin{equation*}
    \Phi(\bx_t) \approx \sum_{n=1}^N a_{nt} \, \Phi(\be_n).
\end{equation*}
Under the nonnegativity of all $\be_N$ and entries of $\bA$, the optimization problem consists in minimizing the sum of the residual errors in the feature space $\cp{H}$, between each $\Phi(\bx_t)$ and its approximation $\sum_{n=1}^N a_{nt} \, \Phi(\be_n)$, namely
\begin{equation}\label{eq:KNMF}
   J_\cp{H}(\bE,\bA)=
   \frac{1}{2} \sum_{t=1}^T \Big\| \Phi(\bx_t) - \sum_{n=1}^N a_{nt} \, \Phi(\be_n) \Big\|_{\cp{H}}^2.
\end{equation}
By analogy to the linear case, a two-block coordinate descent scheme can be investigated to solve this optimization problem.

\section{Input versus feature space optimization}\label{Sec:inputVSfeature}

The difference between the linear and the nonlinear cases is illustrated in \figurename~\ref{fig:bi-NMF}. With the linear NMF, each sample $\bx_t$ is approximated with a linear combination of the $N$ elements $\be_n$, namely by minimizing the Euclidean distance in the input space between each $\bx_t$ and $\widehat{\bx}_t=\sum_{n=1}^N a_{nt} \, \be_n$. With the nonlinear case, using the kernel-based formalism, the optimization is considered in the feature space, by minimizing the distance in $\cp{H}$ between $\Phi(\bx_t)$ and $\widehat{\Psi}_{t}=\sum_{n=1}^N a_{nt} \, \Phi(\be_n)$. The two models, and the corresponding optimization problems, are distinct (except for the trivial linear kernel).

\subsection{The pre-image problem}

An attempt to bridge this gap is to provide a representation of $\widehat{\Psi}_{t}$ in the input space, namely estimating the element of $\cp{X}$ whose image with the mapping function $\Phi(\cdot)$ is as close as possible to $\widehat{\Psi}_{t}$. This is the pre-image problem, which is an ill-posed nonlinear optimization problem; see \cite{11.spm} for more details. As shown in the literature investigating the pre-image problem, and demonstrated recently in \cite[Theorem~1]{13.pr.nn_preimage}, the pre-image takes the form $\sum_{n=1}^N a'_{nt} \, \be_n$, for some unknown coefficients $a'_{nt}$. These coefficients depend on the $\be_n$, making the model implicitly nonlinear.

It is worth noting that this difference, between the linear and the nonlinear case, is inherited from the framework of kernel machines; see \cite{inputVsfeature}. This drawback spans also the multiple kernel learning models, of the form $\sum_{n=1}^N a_{nt} \, \kappa(\be_n,\cdot)$ where the kernel $\kappa$ is a (convex) combination of several kernels \cite{MKL11}. While we focus in this paper on the NMF, our work extends to the wide class of kernel methods, by providing a framework to optimize in both input and feature spaces, as shown next.

\subsection{On augmenting the linear model with a nonlinearity}

Recently, several works have been investigating the combination of a linear model, often advocated by a physical model, with an additive nonlinear fluctuation, determined with a kernel-based term. The model takes the form
\begin{equation*}
	\bx_t = \sum_{n=1}^N a_{nt} \, \be_n + \Psi_{t},
\end{equation*}
where $\Psi_{t}$ belongs to some nonlinear feature space. Several models have been proposed to define this nonlinearity, as outlined here. In \cite{13.tsp.unmix}, the nonlinearity depends exclusively on the endmembers $\be_n$. In \cite{12.whispers.nonlinear}, the above additive fluctuation is relaxed by considering a convex combination with the so-called multiple kernel learning. More recently, the abundances are incorporated in the nonlinear model, with a post-nonlinear model as studied in \cite{13.whispers.postnonlinear} and a Bayesian approach is used in \cite{AltmannDMT14} with the so-called residual component analysis. Another model is proposed in \cite{13.hype.nguyen} in the context of supervised learning.

All these approaches consider that the endmembers $\be_n$ are already known, or estimated using some linear techniques such as the vertex component analysis (VCA) \cite{VCA}. The nonlinearity is only investigated within the abundances $a_{nt}$. As opposed to these approaches, the method considered in this paper investigates also the estimation of the endmembers $\be_n$, with a nonlinear relation with respect to it.

\begin{figure}
\centering
\tiny
\scriptsize
\graphicspath{{Graphics/}}
  \begin{overpic}[width=0.48\textwidth,]{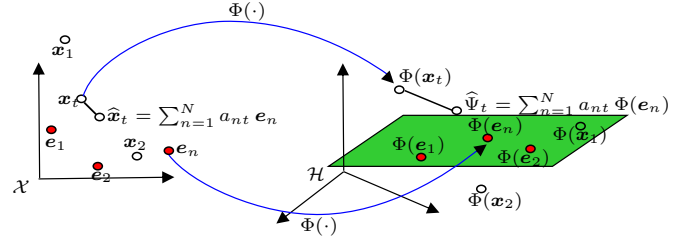}
     \put(9,29){$\bx_1$}
     \put(10,21.5){$\bx_t$}
     \put(17.5,19){$\widehat{\bx}_t=\sum_{n=1}^N a_{nt} \, \be_n$}
     \put(20,15){$\bx_2$}
     \put(8,15){$\be_1$}
     \put(15,10){$\be_2$}
     \put(28,14){$\be_n$}
     \put(62,25){$\Phi(\bx_{t})$}
     \put(85.5,16){$\Phi(\bx_{1})$}
     \put(61,14.5){$\Phi(\be_1)$}
     \put(77,12.5){$\Phi(\be_2)$}
     \put(72.5,17.5){$\Phi(\be_n)$}
     \put(72,20.75){$\widehat{\Psi}_{t}=\sum_{n=1}^N a_{nt} \, \Phi(\be_n)$}
     \put(72.5,6){$\Phi(\bx_2)$}
     \put(36,35){$\Phi(\cdot)$}
     \put(47,2.2){$\Phi(\cdot)$}
     \put(3,8){$\cp{X}$}
     \put(48,10){$\cp{H}$}
  \end{overpic}
  \caption{In the linear NMF, each sample $\bx_t$ is approximated by $\widehat{\bx}_t$ in the input space $\cp{X}$, while in the kernel-based NMF, the mapped sample $\Phi({\bx}_t)$ is approximated by $\widehat{\Psi}_{t}$ in the feature space $\cp{H}$. The proposed bi-objective NMF solves simultaneously the two optimization problems.}
  \label{fig:bi-NMF}
\end{figure}

\section{Bi-objective optimization in both input and feature spaces}\label{Sec:ProposedMethod}


In this section, we propose to solve simultaneously the two optimization problems, in the input and the feature spaces. See \figurename~\ref{fig:bi-NMF} for an illustration.


\subsection{Problem formulation}

Optimizing simultaneously the objective functions $J_\cp{X}(\bE, \bA)$ and $J_\cp{H}(\bE, \bA)$, namely in both the input and the feature space, is in a sense an ill-defined problem. Indeed, it is not possible in general to find a common solution that is optimal for both objective functions. As opposed to the single-objective optimization problems where the main focus would be on the decision solution space, namely the space of all entries $(\bE, \bA)$ (of dimension $LN + NT$), the bi-objective optimization problem brings the focus on the objective space, namely the space to which the objective vector $[J_\cp{X}(\bE, \bA) ~~~ J_\cp{H}(\bE, \bA)]$ belongs. Beyond such bi-objective optimization problem, multi-objective optimization has been widely studied in the literature. Before taking advantage of this literature study and solving our bi-objective optimization problem, we revisit the following definitions in our context:
\begin{itemize}
  \item \textbf{Pareto dominance}: The solution $(\bE_1, \bA_1)$ is said to dominate $(\bE_2, \bA_2)$ if and only if $J_\cp{X}(\bE_1, \bA_1) \leq J_\cp{X}(\bE_2, \bA_2)$ and $J_\cp{H}(\bE_1, \bA_1) \leq J_\cp{H}(\bE_2, \bA_2)$, where at least one inequality is strict.
  \item \textbf{Pareto optimal}: A given solution $(\bE^*, \bA^*)$ is a global (respectively local) Pareto optimal if and only if it is not dominated by any other solution in the decision space (respectively in its neighborhood). That is, the objective vector $[J_\cp{X}(\bE^*, \bA^*) ~~~ J_\cp{H}(\bE^*, \bA^*)]$ corresponding to a Pareto optimal $(\bE^*, \bA^*)$ cannot be improved in any space (input or feature space) without any degradation in the other space.
  \item \textbf{Pareto front}: The set of the objective vectors corresponding to the Pareto optimal solutions forms the Pareto front in the objective space.
\end{itemize}

Various multi-objective optimization techniques have been proposed and successfully applied into engineering fields, {\em e.g.,} the evolutionary algorithms \cite{EvolutionaryAlgo}, sum-weighted algorithms \cite{Das97Drawback,Jong09}, $\varepsilon$-constraint method \cite{ehrgott2008improved,Bérubé200939}, normal boundary intersection method \cite{das1998normal}, to name a few. See the survey \cite{lampinen2000multiobjective,miettinen2008introduction} and the references therein on the methods for multi-objective optimization. Among the existing methods, the sum-weighted or scalarization method has been always the most popular one, since it is straightforward and easily to implement. A sum-weighted technique converts a multi-objective problem into a single-objective problem by combining the multiple objectives. Under some conditions, the objective vector corresponding to latter's optimal solution belongs to the convex part of multi-objective problem's Pareto front. Thus, by changing the weights among the objectives appropriately, the Pareto front of the original problem is approximated. The drawbacks of the sum-weighted method reside in that the nonconvex part of the Pareto front is unattainable, and even on the convex part of the front, a uniform spread of weights does not frequently result in a uniform spread of Pareto points on the Pareto front, as pointed out in \cite{Das97Drawback}. Nevertheless, the sum-weighted method is the most practical one, in view of the complexity of the NMF problem, which is nonconvex, ill-posed and NP-hard.


\subsection{Bi-objective optimization with the sum-weighted method}

Following the formulation introduced in the previous section, we propose to minimize the bi-objective function $[J_\cp{X}(\bE, \bA) ~~~ J_\cp{H}(\bE, \bA)]$, under the nonnegativity of the matrices $\bE$ and $\bA$. The decision solution, of size $LN + NT$, corresponds to the entries in the unknown matrices $\bE$ and $\bA$. In the following, we use the sum-weighted method, which is the most widely-used approach to tackle multi-objective optimization. To this end, we transform the bi-objective problem into an aggregated objective function which is a convex combination of the two original objective functions. Let
\begin{equation*}
	J(\bE, \bA) = \alpha J_{\cp{X}}(\bE, \bA) + (1-\alpha)J_{\cp{H}}(\bE, \bA)
\end{equation*}
be the aggregated objective function ({\em i.e.}, sum-weighted objective function, also called scalarization value) for some weight $\alpha \in [0,1]$ that represents the relative importance between objectives $J_{\cp{X}}$ and $J_\cp{H}$. The optimization problem becomes
\begin{align}\label{eq:optim}
\begin{aligned}
\displaystyle \min_{\bE, \bA}  ~~& \alpha J_{\cp{X}}(\bE, \bA) + (1-\alpha)J_{\cp{H}}(\bE, \bA) \\
\text{subject to} ~~& {\bE} \geq 0 \text{ and } {\bA} \geq 0
\end{aligned}
\end{align}
For a fixed value of the weight $\alpha$, the above problem is called the suboptimization problem. The solution to the above suboptimization problem is a Pareto optimal for the original bi-objective problem, as proven in \cite{Das97Drawback} for the general case. By solving the above suboptimization problem with a spread of values of the weight $\alpha$, we obtain an approximation of the Pareto front. It is obvious that the model breaks down to the single-objective conventional NMF in \eqref{eq:NMF} with $\alpha=1$, while the extreme case with $\alpha=0$ leads to the kernel NMF in \eqref{eq:KNMF}.

The optimization problem \eqref{eq:optim} has no closed-form solution, a drawback inherited from optimization problems with nonnegativity constraints. Moreover, the objective function is nonconvex and nonlinear, making the optimization problem difficult to solve. In the following, we propose iterative techniques for this purpose. It is noteworthy to mention this yields an approximate optimal solution, whose objective vector approximates a point on the Pareto front.
Substituting the expressions given in \eqref{eq:NMF} and \eqref{eq:KNMF} for $J_{\cp{X}}$ and $J_\cp{H}$, the aggregated objective function becomes
\begin{equation}\label{eq:J}
\footnotesize
\begin{split}
J & = \frac{\alpha}{2} \sum_{t=1}^T \Big\| \bx_t - \sum_{n=1}^N a_{nt} \, \be_n \Big\|^2+\frac{1-\alpha}{2} \sum_{t=1}^T \Big\| \Phi(\bx_t) - \sum_{n=1}^N a_{nt} \, \Phi(\be_n) \Big\|_{\cp{H}}^2.
\end{split}
\end{equation}
This objective function takes the form
{\footnotesize
\begin{align}\label{eq:J2}
\begin{aligned}
&\min_{a_{nt}, \be_n}  \alpha\sum_{t=1}^T \Big( -  \sum_{n=1}^N a_{nt}\be_n^\top\bx_t + \frac{1}{2} \sum_{n=1}^N \sum_{m=1}^N a_{nt} a_{mt} \be_n^\top \be_{m} \Big) \\
&+(1-\alpha)\sum_{t=1}^T \Big( -  \sum_{n=1}^N a_{nt} \kappa(\be_n,\bx_t) + \frac{1}{2} \sum_{n=1}^N \sum_{m=1}^N a_{nt} a_{mt} \kappa(\be_n, \be_{m}) \Big),
\end{aligned}
\end{align}
}after expanding the expressions of the distances in \eqref{eq:J}, and removing the constant terms $\bx_t^\top\bx_t$ and $\kappa(\bx_t, \bx_t)$, since they are independent of $a_{nt}$ and $\be_n$.

Although the NMF is nonconvex, its subproblem with one matrix fixed is convex. Similar to most NMF algorithms, we apply the two-block coordinate descent scheme, namely, alternating over the elements in $\bE$ or in $\bA$, while keeping the elements in the other matrix fixed. The derivative of \eqref{eq:J2} with respect to $a_{nt}$ is
\begin{align}\label{eq:derivativeA}
\begin{aligned}
&\nabla\!_{a_{nt}} J = \alpha \Big(-\be_n^\top \bx_t + \sum_{m=1}^N a_{mt} \, \be_n^\top \be_{m}\Big)\\
&+(1-\alpha)\Big(-\kappa(\be_n,\bx_t) + \sum_{m=1}^N a_{mt} \, \kappa(\be_n, \be_{m})\Big),
\end{aligned}
\end{align}
and the gradient of \eqref{eq:J2} with respect to $\be_n$ is
\begin{align}\label{eq:grad_e}
\begin{aligned}
&\nabla\!_{\be_n} J = \alpha \sum_{t=1}^T a_{nt} \Big ( - \bx_t + \sum_{m=1}^N a_{mt}\be_{m}\Big ) \\
&+(1-\alpha)\sum_{t=1}^T a_{nt} \Big ( \!-\! \nabla\!_{\be_n} \kappa(\be_n,\bx_t) \!+ \!\!\sum_{m=1}^N a_{mt} \nabla\!_{\be_n} \kappa(\be_n, \be_{m}) \Big ).
\end{aligned}
\end{align}
Here, $\nabla\!_{\be_n} \kappa(\be_n,\cdot)$ represents the gradient of the kernel with respect to its argument $\be_n$, and can be determined for most valid kernels, as shown in \tablename~\ref{Tab.GradientOfKernels}.

Based on the gradient descent scheme, a simple additive update rule can be written as
\begin{equation}\label{eq:updateaddiA}
  a_{nt} = a_{nt} - \eta_{nt} \, \nabla\!_{a_{nt}} J
\end{equation}
for $a_{nt}$, and
\begin{equation}\label{eq:updateaddiE}
 \be_{n} = \be_{n} - \eta_{n} \nabla\!_{\be_{n}}J
\end{equation}
for $\be_n$.
The stepsize parameters $\eta_{nt}$ and $\eta_{n}$ balance the rate of convergence with the accuracy of optimization, and can be set differently depending on $n$ and $t$. After each iteration, the rectification $a_{nt} = \max (a_{nt}, 0)$ should follow to guarantee the nonnegativity of all $a_{nt}$ and the entries in all $\be_n$.

\subsection{Multiplicative update rules for the Gaussian kernel}
\label{sec:multi}

The additive update rule is easy to implement but the convergence can be slow and very sensitive to the stepsize value, as pointed out by Lee and Seung in \cite{lee00}. Following the spirit of the latter paper, we provide multiplicative update rules for the proposed bi-objective NMF. Without loss of generality, we restrict the presentation to the case of the Gaussian kernel for the second objective function $J_\cp{H}$. For most valid kernels, the corresponding multiplicative update rules can be derived using a similar procedure.

The Gaussian kernel is defined by $\kappa(\bz_i, \bz_j) = \exp (\frac{-1}{2 \sigma^2} \| \bz_i - \bz_j \|^2)$ for any $\bz_i,\bz_j \in \cp{X}$, where $\sigma$ denotes the tunable bandwidth parameter. In this case, its gradient with respect to $\be_n$ is
\begin{equation}\label{eq:GaussianGrad}
\nabla\!_{\be_n} \kappa(\be_n, \bz) = -\frac{1}{\sigma^2} \kappa(\be_n,\bz) (\be_n - \bz).
\end{equation}
To derive the multiplicative update rule for $a_{nt}$, we choose the stepsize parameter in the additive update rule \eqref{eq:updateaddiA} as
\begin{equation*}
	\eta_{nt} = \frac{a_{nt}}{\displaystyle \alpha \sum_{m=1}^N a_{mt}\be_n^\top\be_{m}+(1-\alpha) \sum_{m=1}^N a_{mt} \, \kappa(\be_n, \be_{m})},
\end{equation*}
which yields
\begin{equation}\label{eq:updatemulA}
   a_{nt} = a_{nt} \times \frac{ \alpha \, \be_n^\top\bx_t+(1-\alpha) \,\kappa(\be_n,\bx_t)}{\displaystyle \alpha \sum_{m=1}^N a_{mt}\be_n^\top\be_{m}+ (1-\alpha)\sum_{m=1}^N a_{mt} \, \kappa(\be_n, \be_{m}) }.
\end{equation}
Incorporating the above expression $\nabla\!_{\be_n} \kappa(\be_n, \bz)$ of the Gaussian kernel in \eqref{eq:grad_e}, the gradient of the objective function with respect to $\be_n$ becomes \eqref{eq:GaussianGradE} (given on top of next page). The multiplicative update rule for $\be_n$ is elaborated using the so-called split gradient method \cite{Lanteri11}. This trick decomposes the expression of the gradient~\eqref{eq:grad_e} into the subtraction of two nonnegative terms, {\em i.e.}, $\nabla\!_{\be_n} J = P - Q$, where $P$ and $Q$ have nonnegative entries. To this end, we set the stepsize $\eta_{n}$ corresponding to $\be_n$ in~\eqref{eq:updateaddiE} as~\eqref{eq:StepsizeE} (given on top of next page), and obtain the following multiplicative update for $\be_n$ in~\eqref{eq:updatemulE} (given on top of next page), where the division and multiplication are element-wise.

\begin{table}[t]
\renewcommand{\arraystretch}{1.3}
\centering 
\caption{Some common kernels and their gradients with respect to $\be_n$ }\label{Tab.GradientOfKernels}
\begin{tabular}{|l|c|c|}
\hline
Kernel& $\kappa(\be_n, \bz)$ & $\nabla\!_{\be_n} \kappa(\be_n, \bz)$\\
\Xhline{1pt}
Gaussian&$\exp (\frac{-1}{2 \sigma^2} \| \be_n - \bz\|^2)$    &     $-\frac{1}{\sigma^2} \kappa(\be_n,\bz) (\be_n - \bz)$\\
Polynomial&$(\bz^\top \be_n+ \text{c})^d$     &    $d \,(\bz^\top \be_n+ \text{c})^{(d-1)}\bz$\\
Exponential& $\exp (\frac{-1}{2 \sigma^2} \| \be_n - \bz\|)$   &    $-\frac{1}{2\sigma^2} \kappa(\be_n,\bz) \mathrm{sgn}(\be_n - \bz)$\\
Sigmoid&$\tanh(\gamma\bz^\top \be_n+ \text{c})$   & $\gamma \mathrm{sech}^2(\gamma\bz^\top \be_n+ \text{c})\bz $     \\
\hline
\end{tabular}
\end{table}

\begin{figure*}[!t]
\normalsize
\setcounter{equation}{12}

\begin{equation}\label{eq:GaussianGradE}
\nabla\!_{\be_n} J = \alpha \sum\limits_{t=1}^T a_{nt} \Big ( - \bx_t + \sum\limits_{m=1}^N a_{mt}\be_{m}\Big )
+\frac{1-\alpha}{\sigma^2} \sum\limits_{t=1}^T a_{nt} \Big (\kappa(\be_n,\bx_t) (\be_n - \bx_t)- \sum\limits_{m=1}^N a_{mt} \kappa(\be_n,\be_m)(\be_n - \be_m ) \Big )
\end{equation}

%
\begin{equation}\label{eq:StepsizeE}
\eta_{n} = \frac{\be_{n}}{\alpha \,\sigma^2 \sum\limits_{t=1}^T a_{nt}\sum\limits_{m=1}^N a_{mt}\be_m +(1-\alpha)\sum\limits_{t=1}^T a_{nt}\Big (\kappa(\be_n,\bx_t)\be_n + \sum\limits_{m=1}^N a_{mt} \kappa(\be_n,\be_m)\be_m \Big )  }
\end{equation}

\begin{equation}\label{eq:updatemulE}
    \be_{n} =\displaystyle \be_{n} \otimes \frac{\alpha \sigma^2 \sum\limits_{t=1}^T a_{nt}\bx_t +(1-\alpha)\sum\limits_{t=1}^T a_{nt}\Big ( \kappa(\be_n,\bx_t)\bx_t+ \sum\limits_{m=1}^N a_{mt} \kappa(\be_n,\be_m)\be_n \Big )  }
 {\alpha \, \sigma^2 \sum\limits_{t=1}^T a_{nt}\sum\limits_{m=1}^N a_{mt}\be_m +(1-\alpha)\sum\limits_{t=1}^T a_{nt}\Big (\kappa(\be_n,\bx_t)\be_n
+ \sum\limits_{m=1}^N a_{mt} \kappa(\be_n,\be_m)\be_m \Big )  }
\end{equation}

\hrulefill
\vspace*{4pt}
\end{figure*}

\subsection*{On the convergence, initialization and stopping criteria}

The proposed algorithm tolerates the use of any strictly positive matrices as initial matrices. A simple uniform distribution on the positive quadrant is shown to be a good initialization in our experiments. It is an advantage over some NMF algorithms where stricter initial conditions are required. For instance, proposed for the hyperspectral unmixing problem, both constrained NMF \cite{Jia2009} and minimum volume constrained NMF \cite{EndmemberExtracNMF} initialize the columns of the endmember matrix $\bE$ with randomly chosen pixels from the image under study.

For each given weight $\alpha$, the stopping criterion is two-fold, either a stationary point is attained, or the preset maximum number of iterations is reached. 
Therefore, the algorithm stops at the $n$-th iteration if
\begin{equation*}
  J^{(n)}\leq \min\{J^{(n-1)}, J^{(n+1)}\}
\end{equation*}
or
\begin{equation*}
  n=n_{\max},
\end{equation*}
where $J^{(n)}$ denotes the evaluation of the aggregation objective function $J$ at the $n$-th iteration and $n_{\max}$ is a predefined threshold. 

On the convergence of the proposed algorithm, it is noteworthy to mention that the quotients in the multiplicative update rules \eqref{eq:updatemulA} and \eqref{eq:updatemulE} are unity if and only if
\begin{equation*}
\nabla\!_{a_{nt}} J=0 ~~\text{and}~~ \nabla\!_{\be_n} J=0,
\end{equation*}
respectively. Therefore, the above multiplicative update rules imply a part of the Karush-Kuhn-Tucker (KKT) conditions. 
However, the KKT conditions state only the necessary conditions for a local minimum. Concerning the nonconvex problem as the studied one, or any problem with non-unique KKT-points (stationary points), a local minimum is not guaranteed. Similar to other multiplicative-type update rules proposed in NMF, the proposed algorithm lacks guaranteed optimality property, since the convergence to a stationary point does not always correspond to a local minimum. See also the discussions around the convergence of the conventional NMF in \cite{EFG05Convergence,lin2007projected}. Independently from these theoretical lack of convergence, we show next that the proposed algorithm provides relevant results, and also outperforms all state-of-the-art methods.

\section{Experiments}\label{Sec:Experiments}

In this section, the performance of the proposed algorithm for bi-objective NMF is demonstrated on the unmixing of two well-known hyperspectral images. An approximation of the Pareto front is proposed and a comparison with state-of-the-art unmixing methods is conducted.

\subsection{Dataset and settings}

\begin{figure}[t]\label{Fig.Urban}
\centering
 \graphicspath{{Graphics/}}
\includegraphics[trim = 0mm 0mm 0mm 0mm, clip, width=.48\textwidth]{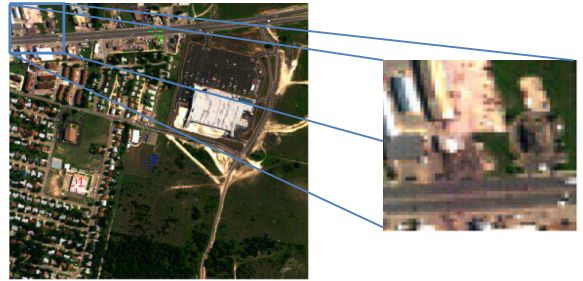}
\caption{The ground truth of the Urban image}\label{Fig.GroundTruthUrban}
\end{figure}

The first image, depicted in \figurename~\ref{Fig.GroundTruthUrban}, is the Urban image\footnote{The Urban image is available: http://www.erdc.usace.army.mil/Media/-\\-FactSheets/FactSheetArticleView/tabid/9254/Article/476681/hypercube.aspx}, acquired by the Hyperspectral Digital Imagery Collection Experiment (HYDICE) sensor. The top left part with $50\times50$ pixels is taken from the original $307\times307$ pixels' image. The raw data consists of 210 channels covering the bandwidth from $0.4\mu m$ to $2.5\mu m$. As recommended in \cite{JiaUrban}, only $L=162$ clean bands of high-SNR are of interest. According to the ground truth provided in \cite{JiaUrban,fong2011hyperactive}, the studied area is mainly composed by grass, tree, building, and road.

The second image is a sub-image with $50\times50$ pixels selected from the well-known Cuprite image, which was acquired by the Airborne Visible/Infrared Imaging Spectrometer (AVIRIS) in 1997. The data is collected over 244 contiguous spectral bands, with the wavelength ranging from $0.4\mu m$ to $2.5\mu m$. After the removal of the noisy bands, $L=189$ spectral bands remain. As investigated in \cite{clark1993mapping,HaAlDoTo2011}, this area is known to be dominated by three materials: muscovite, alunite and cuprite.

Experiments are conducted employing the weight set $\alpha\in\{0,0.02,...,0.98,1\}$, which implies the model varying gradually from the nonlinear Gaussian NMF ($\alpha=0$) to the conventional linear NMF ($\alpha=1$). For each $\alpha$ from the weight set, multiplicative update rules given in \eqref{eq:updatemulA} and \eqref{eq:updatemulE} are applied, with the maximum iteration number $n_{\max}=300$. The initial matrices of $\bE$ and $\bA$ are generated using a [0, 1] uniform distribution. To choose an appropriate bandwidth $\sigma$ in the Gaussian kernel, we first apply the single objective Gaussian NMF on both images, using the same candidate set $\{0.2, 0.3, \ldots, 9.9, 10, 15, 20, \ldots, 50\}$ for $\sigma$. Considering the reconstruction error in both input and feature space (see below for definitions), we fix $\sigma=3.0$ for the Urban image, and $\sigma=2.5$ for the Cuprite image as investigated in \cite{15.tpami.knmf}.

\begin{figure}[t]
\psfragscanon
\psfrag{f}[c][c]{$J$}
\psfrag{X}[c][c]{\scriptsize$\cp{X}$}
\psfrag{H}[c][c]{\scriptsize$\cp{H}$}
\centering
\subfigure[Urban image]{
\label{Fig.ParetoFrontUrban}
 \graphicspath{{Graphics/}}
\includegraphics[trim = 15mm 0mm 0mm 0mm, clip, width=.5\textwidth]{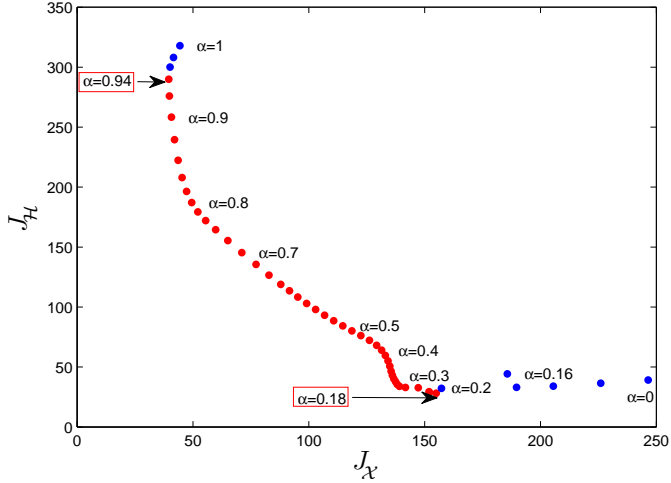}}
\medskip
\subfigure[Cuprite image]{
\label{Fig.ParetoFrontCuprite}
 \graphicspath{{Graphics/}}
\includegraphics[trim = 15mm 0mm 0mm 0mm, clip, width=.5\textwidth]{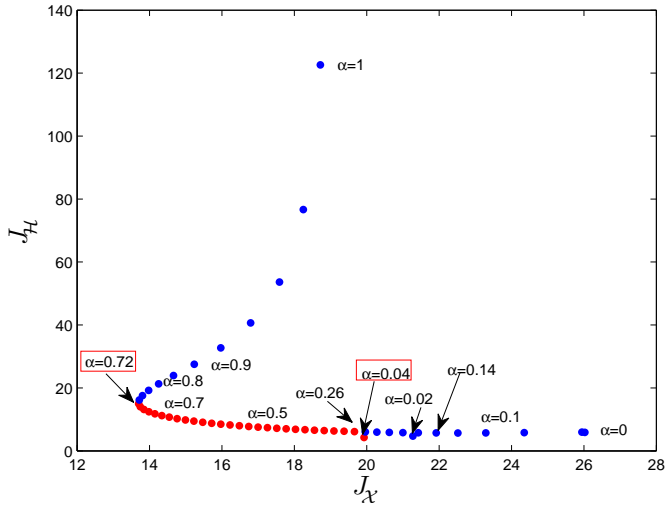}}
\caption{Illustration of the approximated Pareto front in the objective space for Urban and Cuprite images. The objective vectors of the non-dominated solutions (42 for the Urban image, 28 for the Cuprite image), marked in red, approximate a part of the Pareto front; the objective vectors of the dominated solutions (9 for the Urban image, 23 for the Cuprite image) are marked in blue.}
\label{Fig.ParetoFront}
\end{figure}

\subsection{Approximation of the Pareto front}

Since to determine the whole Pareto front is unrealistic for a nonlinear multi-objective optimization problem, one target is to approximate the Pareto front by a set of discrete points on it \cite{lampinen2000multiobjective}. The concept of the Pareto optimal and the Pareto front are not strict in the proposed algorithm, due to the solver of the suboptimization problem not guaranteeing a local minimum, not to mention the global minimum. These obstacles are inherited from the nonconvexity and the nonlinearity of the kernel-based NMF problem. In this case, the Pareto optimal and the Pareto front refer actually to \emph{candidate Pareto optimal} and \emph{an approximation of Pareto front}, respectively \cite{lampinen2000multiobjective}.

\begin{figure}[t]
\psfragscanon
\psfrag{f}[c][c]{$J$}
\psfrag{J}[c][c]{$J$}
\psfrag{X}[c][c]{\scriptsize$\cp{X}$}
\psfrag{H}[c][c]{\scriptsize$\cp{H}$}
\centering
\subfigure[Urban image]{
\label{Fig.ObjectiveFunctionUrban}
 \graphicspath{{Graphics/}}
\includegraphics[trim = 0mm 0mm 0mm 0mm, clip, width=.5\textwidth]{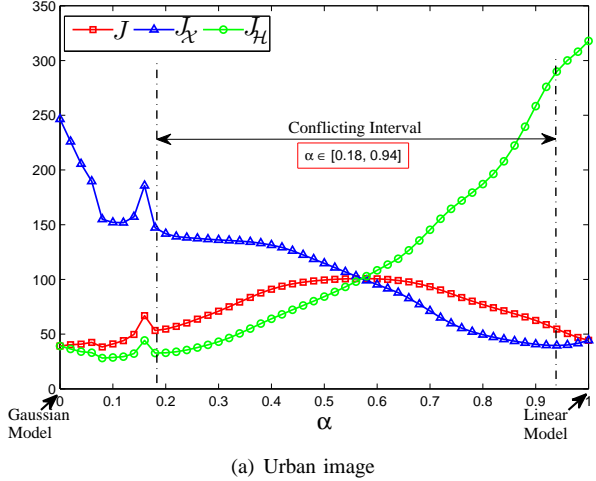}}
\bigskip
\subfigure[Cuprite image]{
\label{Fig.ObjectiveFunctionCuprite}
 \graphicspath{{Graphics/}}
\includegraphics[trim = 0mm 0mm 0mm 0mm, clip, width=.5\textwidth]{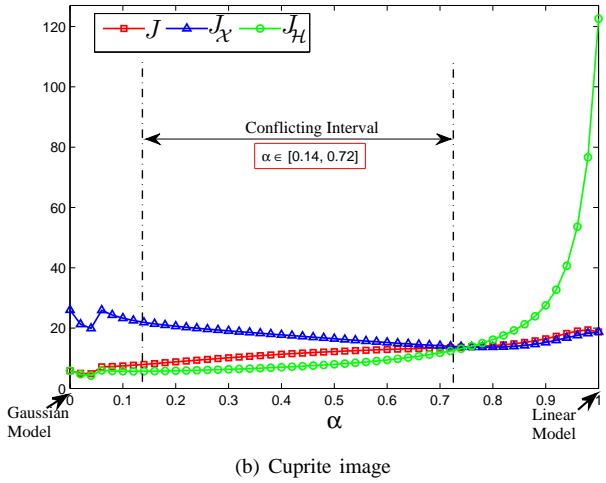}}
\caption{Visualization of the trade-off between the two objectives $J_\cp{X}$ and $J_\cp{H}$, and the change of the aggregated objective function $J$, along with the increment of weight $\alpha$, for the Urban image and the Cuprite image.}
\label{Fig.Objective}
\end{figure}

To approximate the Pareto front with a discrete set of points, we operate as follows: For each value of the weight $\alpha$, we obtain a solution (endmember and abundance matrices) from the algorithm proposed in Section~\ref{sec:multi}; by evaluating the objective functions $J_\cp{X}$ and $J_\cp{H}$ at this solution, we get a single point in the objective space. The approximated Pareto front for the Urban and the Cuprite images are shown in \figurename~\ref{Fig.ParetoFrontUrban} and  \figurename~\ref{Fig.ParetoFrontCuprite}. The evolution of objectives $J_\cp{X}$, $J_\cp{H}$  and the aggregated objective function $J$, evaluated at the solution obtained for each weight $\alpha$, are shown in \figurename~\ref{Fig.ObjectiveFunctionUrban} for the Urban image and in \figurename~\ref{Fig.ObjectiveFunctionCuprite} for the Cuprite image.

We observe the following:
\begin{enumerate}[1)]
  \item For both images under study, solutions generated with $\alpha=1$ and $\alpha=0$ are dominated, since all the solutions on the Pareto front outperform them, with respect to both objectives. This reveals that neither the conventional linear NMF nor the nonlinear Gaussian NMF best fits the studied images. On the contrary, the Pareto optimal solutions, which result the points on the Pareto front, provide a set of feasible and nondominated decompositions for the decision maker (DM), {\em i.e.}, the user. It is worth noting that we apply the sum-weighted method as a \emph{posteriori} method, where different Pareto optimal solutions are generated, and the DM makes the final comprise among optimal solutions. Alternatively, in a \emph{priori} method, the DM specifies the weight $\alpha$ in advance to generate a solution. See \cite{miettinen2008introduction} for more details.

  \item Regarding the sum-weighted approach, the minimizer of the suboptimization problem is proven to be a Pareto optimal for the original multi-objective problem, {\emph{i.e.}}, the corresponding objective vector belongs to the Pareto front in the objective space \cite{Das97Drawback}. In practice, we obtain 9 and 23 (out of 51) dominated solutions for the Urban and the Cuprite images, respectively. Such phenomenon, however, is not surprising, since \emph{there exist multiple Pareto optimal solutions in a problem only if the objectives are conflicting to each other}, as claimed in \cite{MultiDeb01}\footnote{For example, the Pareto optimal solutions for the well-known Schaffer's function, defined by $J(x)=[x^2,(x-2)^2]^\top$, are found only within the interval $[0, 2]$, where a tradeoff between two objectives exists. See \cite{EvolutionaryAlgo} for more details.}. Other possible explanation could be the applied numerical optimization scheme, due to the weak convergence of the method or to the failure of the solver in finding a global minimum \cite{miettinen2008introduction}. For the Urban image as shown in \figurename~\ref{Fig.ParetoFrontUrban} and \figurename~\ref{Fig.ObjectiveFunctionUrban}, all the obtained solutions are Pareto optimal within the objectives-conflicting interval $\alpha \in [0.18, 0.94]$. Regarding the Cuprite image, as observed in \figurename~\ref{Fig.ParetoFrontCuprite} and \figurename~\ref{Fig.ObjectiveFunctionCuprite}, the objectives-conflicting interval is $\alpha\in [0.14, 0.72]$, while the Pareto optimal solutions are found using $\alpha\in \{0.04\}\cup \{0.26, 0.28, ..., 0.72\}$. In fact, the obtained solutions with $\alpha\in \{0.14, 0.16, ..., 0.24\}$ are only local Pareto optimal, and they are dominated by a global Pareto optimal with $\alpha=0.04$. It is pointed out in \cite{miettinen2008introduction} that, in the nonconvex problem, the global (local) solver generates global (local) Pareto optimal, and local Pareto optimal is not of interest in front of global Pareto optimal.
  \item As illustrated in both \figurename~\ref{Fig.ParetoFrontUrban} and \figurename~\ref{Fig.ParetoFrontCuprite}, an even distribution of weight $\alpha$ between [0, 1] do not lead to an even spread of the solutions on the approximated Pareto front. Moreover, the nonconvex part of the Pareto front cannot be attained using any weight. It is exactly the case in \figurename~\ref{Fig.ParetoFrontCuprite}; in \figurename~\ref{Fig.ParetoFrontUrban}, a trivial nonconvex part between $\alpha=0.3$ and $\alpha=0.5$ on the approximated Pareto front is probably resulted from the nonoptimal solution of the suboptimization problem. These are two main drawbacks of the sum-weighted method.
\end{enumerate}

Nevertheless, the obtained approximation of Pareto front is of high value. On one hand, it provides a set of Pareto optimal solutions for the DM, instead of a single decomposition. On the other hand, an insight of the trade-off between objectives $J_\cp{X}$ and $J_\cp{H}$  reveals the underlying linearity/nonlinearity of the data under study, as illustrated in the following section.

\subsection{Performance}

In this section, we study the performance of the method on the unmixing problem in hyperspectral imagery. The unmixing performance is evaluated by two metrics introduced in \cite{15.tpami.knmf}. The first one is the reconstruction error in the input space (RE) defined by
\begin{align*}
   \text{RE}=&\sqrt{\frac{1}{TL}\sum_{t=1}^T \|\bx_{t}-{\sum_{n=1}^{N}a_{nt}\be_{t}}\|^2}. \notag
\end{align*}
The second one is the reconstruction error in the feature space ($\text{RE}^\Phi$), which is similarly defined as
\begin{equation*}
     \text{RE}^\Phi=\sqrt{\frac{1}{TL}\sum_{t=1}^T \Big\|\Phi (\bx_{t})-{\sum_{n=1}^{N}a_{nt}\Phi(\be_{t})} \Big\|_\cp{H}^2}, \notag
\end{equation*}
where
\begin{align*}
  \Big\|\Phi (\bx_{t}) - \sum_{n=1}^{N}a_{nt}\Phi(\be_{t}) \Big\|_\cp{H}^2
  = &\sum_{n=1}^N \sum_{m=1}^N a_{nt} a_{mt} \kappa(\be_n, \be_{m})
  \\
  &-2\sum_{n=1}^N a_{nt} \kappa(\be_n,\bx_t)
	+ \kappa(\bx_t,\bx_t),
\end{align*}
and $\kappa(\cdot,\cdot)$ denotes the Gaussian kernel. It is worth to note that $\text{RE}^\Phi$ can always be evaluated for any given matrices $\bE$ and $\bA$, regardless of the optimization problem and the solving procedure that led to these matrices.

\subsection*{State-of-the-art unmixing methods}

An unmixing problem comprises the estimation of endmembers and the corresponding abundance maps. Some existing techniques either extract the endmembers (such as VCA) or estimate the abundances (such as FCLS)\footnote{See \cite{12.tgrs.barycenters} for connections between the endmember extraction techniques and the abundances estimation techniques.}; other methods enable the simultaneous estimations, \emph{e.g.,} NMF and its variants. We briefly present all the unmixing algorithms that are used in comparison.

The most-known endmember extraction technique is the vertex component analysis (VCA) \cite{VCA}. It is based on the linear mixture model and presumes the existence of endmembers within the image under analysis. It seeks to inflate the simplex enclosing all the spectra. The endmembers are the vertices of the largest simplex. This technique is applied for endmember extraction, jointly with three abundance estimation techniques: FCLS, K-Hype and GBM-sNMF. The fully constrained least squares algorithm (FCLS) \cite{Heinz} is a least square approach using the linear mixture model, where the abundances are estimated considering the nonnegativity and sum-to-one constraints. A nonlinear unmixing model for abundance estimation is considered in \cite{13.tsp.unmix}, where the nonlinear term is described as a kernel-based model, with the so-called linear-mixture/nonlinear-fluctuation model (K-Hype). In \cite{Yokoya14}, a generalized bilinear model is formulated, with parameters optimized using the semi-nonnegative matrix factorization (GBM-sNMF).

We further consider five NMF-based techniques that are capable to estimate the endmembers and abundances jointly. The minimum dispersion constrained NMF (MinDisCo) \cite{Huck} includes the dispersion regularization to the conventional NMF, by integrating the sum-to-one constraint for each pixel's abundance fractions and the minimization of variance within each endmember. The problem is solved by exploiting an alternate projected gradient scheme. In the convex nonnegative matrix factorization (ConvexNMF) \cite{Ding10}, the basic matrix (endmember matrix in our context) is restricted to the span of the input data, that is, each sample can be viewed as a convex combination of certain data points. The kernel convex-NMF (KconvexNMF) and the kernel semi-NMF based on the nonnegative least squares (KsNMF) are essentially the kernelized variants of the ConvexNMF in \cite{Ding10} and the alternating nonnegativity constrainted least squares and the active set method in \cite{Kim2008}, respectively, as discussed in \cite{Li2012}. Experiments are also conducted with these two kernel methods, adopting the Gaussian kernel.
Nonlinear NMF based on constructing Mercer kernels (MercerNMF), introduced in \cite{MercerNMF}, addresses the nonlinear NMF problem using a self-constructed Gaussian kernel, where the nonnegativity of the embedded bases and coefficients is preserved. The embedded data are finally factorized with conventional NMF. Of particular note is that only the reconstruction error in the feature space can be calculated for the aforementioned kernel-based methods, since the pre-images of the mapped endmember, which are required in the computing of reconstruction error in the input space, cannot be exploited.

\begin{table}[t]
\renewcommand{\arraystretch}{1.3}
\centering 
\caption{Unmixing performance for the Urban image}\label{Tab.UnmixingUrban}
\begin{tabular}{@{}|@{\;}c@{\,}|l|c|c|c|}
\cline{4-5} 
\multicolumn{3}{c|}{}&RE \tiny$\times {10}^{-2}$\! & $\text{RE}^\Phi$\tiny$\times {10}^{-2}$\!  \\
\cline{2-5}
\multicolumn{1}{c|}{}&\multicolumn{2}{c|}{FCLS}&1.44&3.89\\ 
\cline{2-5}
\multicolumn{1}{c|}{}&\multicolumn{2}{c|}{GBM-sNMF}&6.50&4.11\\
\cline{2-5}
\multicolumn{1}{c|}{}&\multicolumn{2}{c|}{K-Hype}&5.99&4.67\\
\cline{2-5}
\multicolumn{1}{c|}{}&\multicolumn{2}{c|}{MinDisCo}&3.12&4.60\\
\cline{2-5}
\multicolumn{1}{c|}{}&\multicolumn{2}{c|}{ConvexNMF}&2.96&5.84\\
\cline{2-5}
\multicolumn{1}{c|}{}&\multicolumn{2}{c|}{KconvexNMF}&-&43.94\\
\cline{2-5}
\multicolumn{1}{c|}{}&\multicolumn{2}{c|}{KsNMF}&-&4.33\\
\cline{2-5}
\multicolumn{1}{c|}{}&\multicolumn{2}{c|}{MercerNMF}&-&2.96\\
\hline
\multirow{5}{*}{\rotatebox{90}{ this paper\;}} & {LinearNMF}& $\alpha=1$&1.48&3.96\\
\cline{2-5}
&{GaussianNMF}&$\alpha=0$&3.49&1.39\\
\cline{2-5}
&\multirow{3}{*}{Pareto Optimal} &$\alpha=0.18$&2.70 &\textbf{1.27}\\
\cline{3-5}
&&$\alpha=0.50$&2.38 &2.04\\
\cline{3-5}
&&$\alpha=0.94$&\textbf{1.40}&3.78\\
\hline
\end{tabular}
\end{table}

\begin{table}[t]
\renewcommand{\arraystretch}{1.3}
\centering 
\caption{Unmixing performance for the Cuprite image}\label{Tab.UnmixingCuprite}
\begin{tabular}{@{}|@{\;}c@{\,}|l|c|c|c|}
\cline{4-5} 
\multicolumn{3}{c|}{}&RE \tiny$\times {10}^{-2}$\! & $\text{RE}^\Phi$\tiny$\times {10}^{-2}$\!  \\
\cline{2-5}
\multicolumn{1}{c|}{}&\multicolumn{2}{c|}{FCLS}&0.95&0.59\\ 
\cline{2-5}
\multicolumn{1}{c|}{}&\multicolumn{2}{c|}{GBM-sNMF}&1.06&0.62\\
\cline{2-5}
\multicolumn{1}{c|}{}&\multicolumn{2}{c|}{K-Hype}&2.12&0.93\\
\cline{2-5}
\multicolumn{1}{c|}{}&\multicolumn{2}{c|}{MinDisCo}&1.62&4.54\\
\cline{2-5}
\multicolumn{1}{c|}{}&\multicolumn{2}{c|}{ConvexNMF}&1.61&2.51\\
\cline{2-5}
\multicolumn{1}{c|}{}&\multicolumn{2}{c|}{KconvexNMF}&-&19.53\\
\cline{2-5}
\multicolumn{1}{c|}{}&\multicolumn{2}{c|}{KsNMF}&-&1.38\\
\cline{2-5}
\multicolumn{1}{c|}{}&\multicolumn{2}{c|}{MercerNMF}&-&2.74\\
\hline
\multirow{5}{*}{\rotatebox{90}{ this paper\;}} & {LinearNMF}& $\alpha=1$&0.89&2.28\\
\cline{2-5}
&{GaussianNMF}&$\alpha=0$&1.05&0.50\\
\cline{2-5}
&\multirow{3}{*}{ParetoOptimal} &$\alpha=0.04$&0.92 &\textbf{0.42}\\
\cline{3-5}
&&$\alpha=0.50$&0.84 &0.58\\
\cline{3-5}
&&$\alpha=0.72$&\textbf{0.77}&0.73\\
\hline
\end{tabular}
\end{table}

\begin{figure}
\centering
\graphicspath{{Graphics/Urban/}}
\begin{minipage}{0.9\linewidth}
\centering
\subfigure{
\includegraphics[trim = 40mm 60mm 35mm 50mm, clip,width=1\textwidth]{End_Alpha1.eps}}
\subfigure{
\includegraphics[trim = 40mm 65mm 35mm 56mm, clip,width=1\textwidth]{Abun_Alpha1.eps}}
\centerline{(a) $\alpha=1$ }
\end {minipage}
\vfill
\begin{minipage}{1\linewidth}
\centering
\subfigure{
\includegraphics[trim = 40mm 60mm 35mm 50mm, clip,width=0.9\textwidth]{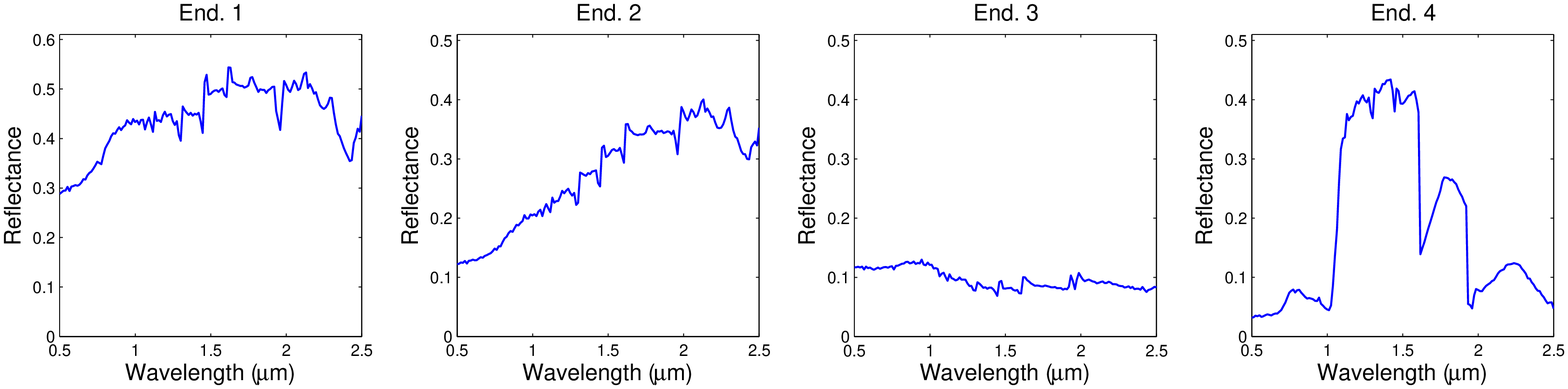}}
\subfigure{
\includegraphics[trim = 40mm 65mm 35mm 56mm, clip,width=1\textwidth]{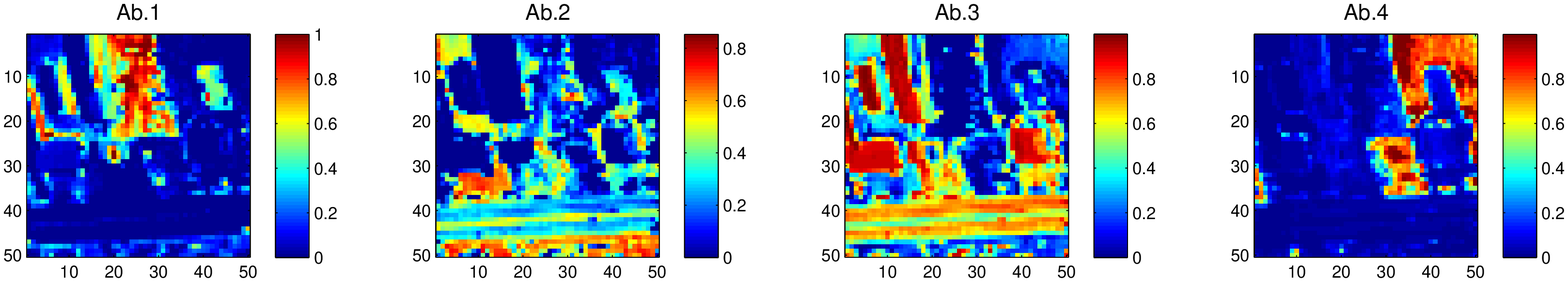}}
\centerline{(b) $\alpha=0.3$  }
\end {minipage}
\vfill
\begin{minipage}{1\linewidth}
\centering\emph{}
\subfigure{
\includegraphics[trim = 40mm 60mm 35mm 50mm, clip,width=1\textwidth]{End_Alpha0.eps}}
\subfigure{
\includegraphics[trim = 40mm 65mm 35mm 56mm, clip,width=1\textwidth]{Abun_Alpha0.eps}}
\centerline{(c) $\alpha=0$ }
\end {minipage}
\caption{Urban image: Endmembers and corresponding abundance maps, estimated using  $\alpha=1$ (conventional linear NMF); $\alpha=0.3$ (a Pareto optimal of the bi-objective NMF); $\alpha=0$ (nonlinear Gaussian NMF).}
\label{Fig. EndAbunUrban}
\end{figure}

\begin{figure}
\centering
\graphicspath{{Graphics/Cuprite/}}
\begin{minipage}{0.9\linewidth}
\centering
\subfigure{
\includegraphics[trim =21mm 22mm 21mm 21mm, clip,width=1\textwidth]{End_Alpha1.eps}}
\subfigure{
\includegraphics[trim = 21mm 28mm 21mm 25mm, clip,width=1\textwidth]{Abun_Alpha1.eps}}
\centerline{(a) $\alpha=1$ }
\end {minipage}
\vfill
\begin{minipage}{0.9\linewidth}
\centering
\subfigure{
\includegraphics[trim = 21mm 22mm 21mm 21mm, clip,width=1\textwidth]{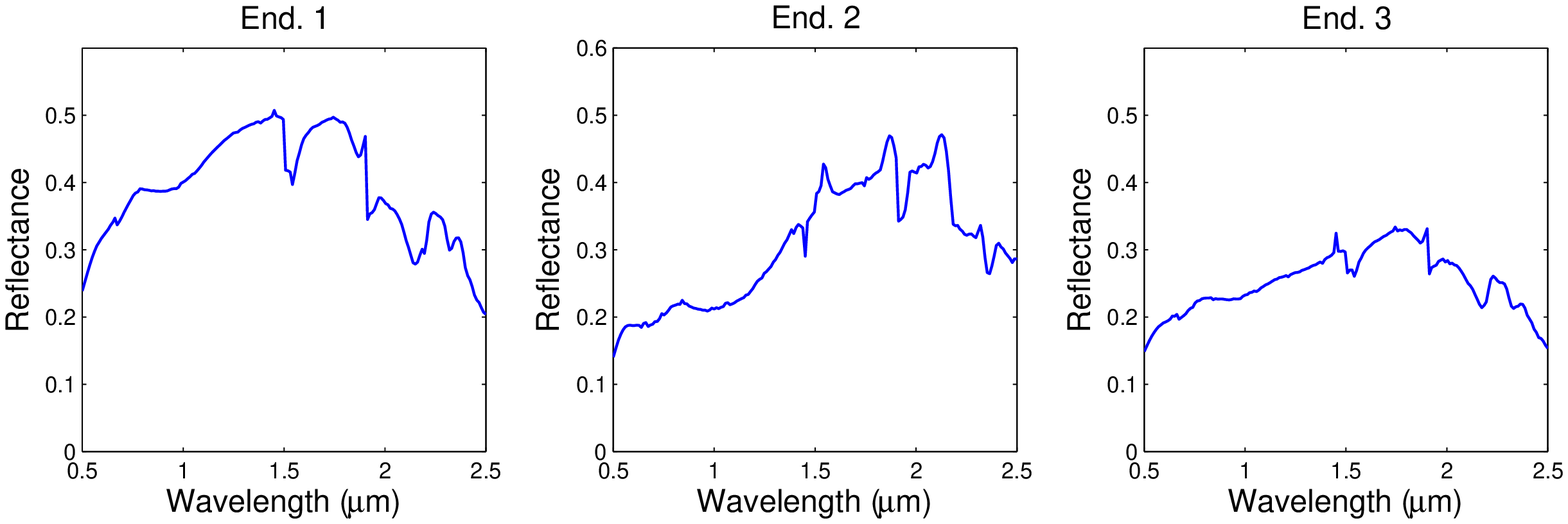}}
\subfigure{
\includegraphics[trim = 21mm 28mm 21mm 25mm, clip,width=1\textwidth]{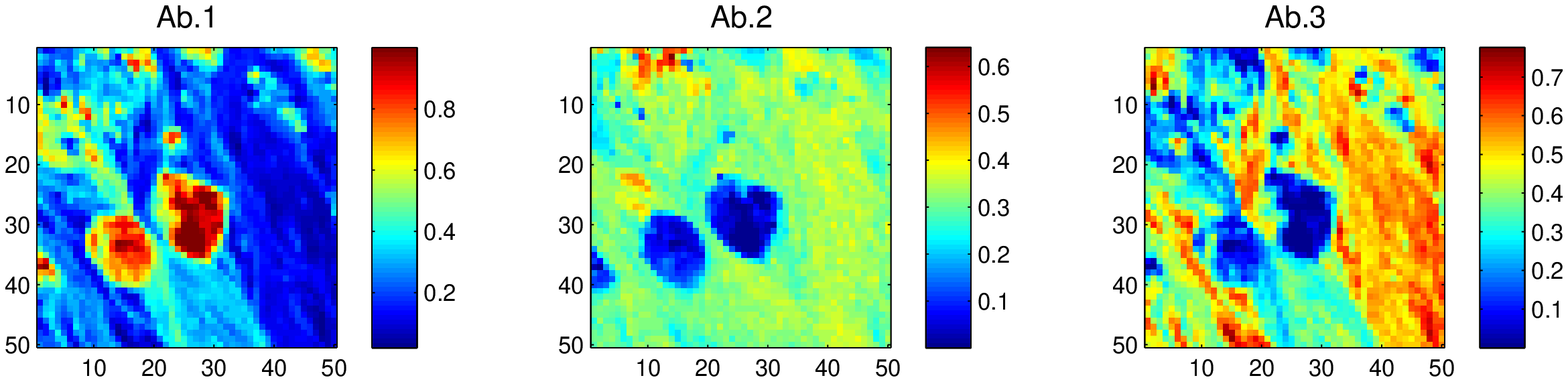}}
\centerline{(b) $\alpha=0.72$  }
\end {minipage}
\vfill
\begin{minipage}{0.9\linewidth}
\centering\emph{}
\subfigure{
\includegraphics[trim =21mm 22mm 21mm 21mm, clip,width=1\textwidth]{End_Alpha0.eps}}
\subfigure{
\includegraphics[trim = 21mm 28mm 21mm 25mm, clip,width=1\textwidth]{Abun_Alpha0.eps}}
\centerline{(c) $\alpha=0$ }
\end {minipage}
\caption{Cuprite image: Endmembers and corresponding abundance maps, estimated using  $\alpha=1$ (conventional linear NMF); $\alpha=0.72$ (a Pareto optimal of the bi-objective NMF); $\alpha=0$ (nonlinear Gaussian NMF).}
\label{Fig. EndAbunCuprite}
\end{figure}

\subsection*{Unmixing performance}

The unmixing performance, with respect to the reconstruction errors in the input and the feature spaces, is compared with the aforementioned unmixing approaches, as demonstrated in \tablename~\ref{Tab.UnmixingUrban} and \tablename~\ref{Tab.UnmixingCuprite}. As can be observed, the proposed method with Pareto optimal solution outperforms not only the existing linear NMF ($\alpha =1$) and Gaussian NMF ($\alpha =0$), but also all the state-of-the-art methods.

The estimated endmembers and the corresponding abundance maps with the proposed method are shown in \figurename~\ref{Fig. EndAbunUrban} and \figurename~\ref{Fig. EndAbunCuprite}. For the Urban image, different areas (the road in particular) are better recognized with the Pareto optimal when compared with solutions of the linear and of the Gaussian NMF. Regarding the Cuprite image, the linear NMF ({\em i.e.}, $\alpha =1$) recognizes two out of three regions; whereas both the Pareto optimal ({\em i.e.}, $\alpha =0.72$) and the Gaussian NMF ({\em i.e.}, $\alpha =0$) are able to distinguish three regions. However, the abundance maps of Gaussian NMF appear to be overly sparse, compared with its counterpart of the Pareto optimal solution. It is also noticed that the endmembers extracted with the linear NMF are spiky, and even with some zero-parts, thus meeting poorly the real situation.

\section{Conclusion}\label{Sec:Conclusion}

This paper presented a novel bi-objective nonnegative matrix factorization by exploiting the kernel machines, where the decomposition was performed simultaneously in the input and the feature space. The multiplicative update rules were derived. The performance of the method was demonstrated for unmixing well-known hyperspectral images. The resulting Pareto fronts were analyzed. As for future work, we are extending this approach to include other NMF objective functions, defined in the input or the feature space. Considering simultaneously several kernels, and as a consequence several feature spaces, is also under investigation.


%

%

\section*{Acknowledgment}

This work was supported by the French ANR, grant \mbox{HYPANEMA: ANR-12BS03-0033}.

\ifCLASSOPTIONcaptionsoff
  \newpage
\fi



%
%
%

\bibliographystyle{IEEEtran}
\bibliography{bib_fei,biblio_ph}
%

\begin{IEEEbiography}{Fei Zhu}
was born in Liaoning, China, in 1988. She received the B.S degrees in mathematics and applied mathematics and in economics in 2011 from the Xi'an Jiaotong University, Xi'an, and M.S degree in systems optimization and security in 2013 from the University of Technology of Troyes (UTT), Troyes, France. She is currently working toward the Ph.D. degree with the University of Technology of Troyes (UTT). Her research interests include hyperspectral image analysis.
\end{IEEEbiography}

\begin{IEEEbiography}{Paul Honeine}
(M'07) was born in Beirut, Lebanon, on October 2, 1977. He received the Dipl.-Ing. degree in mechanical engineering in 2002 and the M.Sc. degree in industrial control in 2003, both from the Faculty of Engineering, the Lebanese University, Lebanon. In 2007, he received the Ph.D. degree in Systems Optimisation and Security from the University of Technology of Troyes, France, and was a Postdoctoral Research associate with the Systems Modeling and Dependability Laboratory, from 2007 to 2008. Since September 2008, he has been an assistant Professor at the University of Technology of Troyes, France. His research interests include nonstationary signal analysis and classification, nonlinear and statistical signal processing, sparse representations, machine learning. Of particular interest are applications to (wireless) sensor networks, biomedical signal processing, hyperspectral imagery and nonlinear adaptive system identification. He is the co-author (with C. Richard) of the 2009 Best Paper Award at the IEEE Workshop on Machine Learning for Signal Processing. Over the past 5 years, he has published more than 100 peer-reviewed papers.
\end{IEEEbiography}



\end{document}